# Learning Generalized Non-Rigid Multimodal Biomedical Image Registration from Generic Point Set Data

Zachary MC Baum[1,2,3], Tamas Ungi[3,4], Christopher Schlenger[3], Yipeng Hu[1,2], Dean C Barratt[1,2]

[1] Centre for Medical Image Computing, University College London
[2] Wellcome/EPSRC Centre for Surgical & Interventional Sciences, University College London
[3] Verdure Imaging, Inc.
[4] Laboratory for Percutaneous Surgery, School of Computing, Queen's University
`zachary.baum.19@ucl.ac.uk`

**Abstract.** Free Point Transformer (FPT) has been proposed as a data-driven, non-rigid point set registration approach using deep neural networks. As FPT does not assume constraints based on point vicinity or correspondence, it may be trained simply and in a flexible manner by minimizing an unsupervised loss based on the Chamfer Distance. This makes FPT amenable to real-world medical imaging applications where ground-truth deformations may be infeasible to obtain, or in scenarios where only a varying degree of completeness in the point sets to be aligned is available. To test the limit of the correspondence finding ability of FPT and its dependency on training data sets, this work explores the generalizability of the FPT from well-curated non-medical data sets to medical imaging data sets. First, we train FPT on the ModelNet40 dataset to demonstrate its effectiveness and the superior registration performance of FPT over iterative and learning-based point set registration methods. Second, we demonstrate superior performance in rigid and non-rigid registration and robustness to missing data. Last, we highlight the interesting generalizability of the ModelNet-trained FPT by registering reconstructed freehand ultrasound scans of the spine and generic spine models without additional training, whereby the average difference to the ground truth curvatures is 1.3°, across 13 patients.

## 1 Introduction

Registration is a fundamental problem wherein corresponding features of images, point sets, or other data that define two objects are transformed spatially so that they are aligned. Registration algorithms determine the rigid or non-rigid transformation which aligns a source object to a target object.

In point set registration, iterative optimization minimizes alignment metrics or cost functions which quantify the misalignment between the source and target points. Usually, methods are unable to handle large point sets in real-time owing to their computationally intense nature. In contrast, the efficient inference and the ability to model complex, non-linear transformations of learning-based methods have seen the development of neural networks which infer the optimal transformation [1-7].

Non-rigid registration is important in image-guided interventions as it permits the fusion of imaging with spatial information and enables soft tissue motion compensation. Models which integrate local deformations have demonstrated improved registration in soft tissues by compensating for patient motion and other anatomical deformations [8]. However, these methods often rely on constrained models or need to explicitly model noise, outliers, and missing data. As such, they may be inadequate to handle real-world data. An additional problem, partial registration, is a common real-world application of registration methods where part of either or both input point sets are missing. This is often the case in the registration of point sets or images from different modalities or points in time as there may not be a one-to-one correspondence between the data, requiring a series of consecutive components to be used to create a global alignment.

While iterative non-rigid point set registration methods exist [5, 9-13], few are learning-based [5, 12-13]. One method which has been applied successfully is the Free Point Transformer (FPT) [12-13]. Importantly, FPT is not limited by the inherently unordered structure of point sets and requires no iterative inference or refinement process, unlike many existing methods [1-4]. As such, FPT could directly perform intra-operative prostate biopsy guidance via point set registration of prostate glands extracted from paired magnetic resonance and transrectal ultrasound imaging [12-13].

In this work, we demonstrate the generalizability of FPT for point set registration. First, we demonstrate robustness to deformation and missing data on the ModelNet40 dataset. Second, using FPT's model-free approach and data-driven learning process, we apply a pre-trained FPT to real-world medical imaging data from an external domain, registering 3D reconstructions of segmented ultrasound (US) scans and a generic spine atlas. In this real-world use case for non-rigid point set registration, we investigate FPT's effectiveness to quantify spinal curvature for scoliosis measurement. Scoliosis is a spinal deformity identified in 3% of adolescent children [14]. It is often monitored and measured with X-ray imaging (X-ray), however, repeated use of X-rays has been linked to an increased incidence of cancer [15]. US has been proposed as a safer, more accessible option for scoliosis monitoring and measurement [16-18], and deep learning methods for automatic US bone segmentation have been shown to adequately reconstruct the spinal curvature in pediatric patients with scoliosis [18]. However, compared to X-rays, reconstructions alone are not sufficient. The registration of such reconstructions to a generic spine model may provide appropriate and meaningful visualizations. Manual registration of such reconstructions to generic spine models for visualization and measurement is possible, though the process is time-consuming, error-prone, and operator-dependent [17]. These limitations reveal the need for fast, automatic methods which register US-based reconstructions to generic anatomical models for use in place of or to supplement traditional medical imaging.

## 2 Methods

### 2.1 Network Architecture

The FPT network architecture is comprised of two modules: a global feature extractor, and a point transformer [13]. To perform the feature extraction, FPT utilizes a modified version of PointNet [19], a neural network architecture previously proposed by Qi *et al.* for converting point sets into permutation and rotation invariant feature vectors for classification and segmentation. These modifications ensure the translation is captured in the feature vectors, primarily through the removal of Batch Normalization [13].

In FPT, twin, weight-sharing PointNets generate feature vectors from each of the source and the target point sets. These are concatenated into a global feature vector, paired with the source points, and fed into a multi-layer perceptron (MLP). The MLP yields a per-point estimation of the unique displacement vector that transforms each point of the source point set without any smoothness or coherence constraints.

### 2.2 Implementation and Training Details

FPT was trained on the ModelNet40 dataset, consisting of 12,311 geometric surface models of 40 object categories, split into a 9,843 object training set and 2,468 object testing set [20]. Input point sets comprised 2048 points sampled from the surfaces of ModelNet40 shapes and were used as target and source point sets in training and testing. Our implementation of FPT and the hyperparameters described below are consistent with the original defaults, as described in [13].

We apply an unsupervised training strategy. This requires the indirect computation of a distance metric between point sets without a known bijective correspondence. Therefore, we utilized the Chamfer Distance [21], as it suits the registration problem well, among distance metrics or losses that do not require bijective correspondence. FPT was trained with the Adam optimizer [22], a minibatch size of 32, and a learning rate of 0.001. Models were trained on a NVIDIA DGX-1 system using a single Tesla V100 GPU. During training, points were transformed on-the-fly with scaling, deformation, rotation, and displacement. Points were scaled, per-sample, between [-1, 1] in the X, Y and Z directions. The scaled points were used as the target point set. Rotations were sampled from [-45°, 45°] about the X, Y, and Z axes. Displacements were sampled from [-1, 1] in the X, Y, and Z directions. Non-rigid deformation was simulated by a radial basis function (RBF) deformation model with a Gaussian kernel. RBF deformation was defined by a perturbation of the control points by Gaussian random shift ($\mu = 0$, $\sigma = 0.1$). These points were used as the source point set. A RBF was chosen as our deformation model as it produces smooth deformations and is sufficiently computationally efficient to apply on-the-fly during training. Experiments with other deformation models, such as elastic body spline models [23], are needed to assess FPT's ability to reconstruct more localized displacements.

When training FPT for partial registration, occlusion occurred after scaling, and was simulated by selecting a random point on the model surface, and discarding the 512

(25% of the input points) nearest neighbour points. Following point removal, translation, rotation, and deformation were performed as described above.

## 3  Experiments

We present FPT's performance in rigid and non-rigid registration and scenarios with partially occluded data. In these experiments, we used the 2,468 object ModelNet40 testing set to compare FPT with existing and classical and learning-based methods. Furthermore, we demonstrate how a pre-trained FPT model may generalize to data outside its training domain in an application whereby segmented US imaging and a generic spine model are non-rigidly registered. Root mean square error (RMSE) is used to evaluate errors in rotation (R) and translation (t). Inference time is also reported.

### 3.1  Registration on ModelNet40

**Unseen Objects**. We evaluated FPT on rigid and non-rigid transformations. Transformations are generated as in training, although for the rigid transformations, no non-rigid deformation was added. As such, the inherent design and training of FPT is unchanged – meaning it may predict a transformation which contains deformation to the input point sets, but the ground truth transform between the input point sets does not have any deformation added for the rigid transformation registrations. We compared FPT to rigid iterative and learning-based methods, as well as non-rigid iterative methods.

**Partial Registration with Unseen Objects**. We evaluated the performance of FPT for partial registration. Transformations are generated as in training for the 'partial-to-full' registrations, where point removal is only performed on the source point set. For 'partial-to-partial' registrations, point removal is performed on both the source and target point sets. We compare FPT to rigid iterative and learning-based methods. Notably, one such method [4] is designed explicitly for partial registration, whereas FPT is not. In this comparison, we note that we modify only the data, as in Section 2.2, and not the FPT architecture.

### 3.2  3D Spine Ultrasound Reconstruction Registration

We evaluate FPT for a medical image registration task. Using Ungi *et al.*'s method for automatic bone segmentation from US [18], 13 different curvatures from 7 different patients were reconstructed. The surfaces of the reconstructions and a generic spine model were resampled into point sets, and subsequently used to evaluate FPT's ability to register the two. As discussed previously, this task is well-suited to a non-rigid partial point set registration method given the inherent differences in the appearance and geometry of the 3D US reconstructions and the generic spine models (Figure 1) due to the fact that the reconstruction may feature partial occlusions due to segmentation quality or operator error during the scanning process. As such, to demonstrate FPT's generalizability to unseen objects from outside the domain of our training set and FPT's ability to perform partial registration; we use the pre-trained FPT model trained for the experiments described in Section 3.1 for 'Partial Registration with Unseen Objects'.

Spinal curvature is commonly quantified by the Cobb angle; the angle between the end-plates of vertebrae above and below the main curvature [25]. However, as vertebral end-plates are not visible in US, they do not appear in the 3D reconstructions. As such, we report spinal curvature using the transverse process angle (TxA) as it is visible in X-ray and US, and has a very strong correlation to Cobb angle [16]. TxA is defined by the angle between the lateral ends of each transverse process above and below the main curvature [16]. TxAs were calculated on the X-ray and the deformed spine model. The reported TxA in the X-ray was defined by two lines in 2D. The TxA in the deformed model was defined by two lines in 3D, with the reported TxA being that which was computed by projecting the lines into 2D in the coronal plane.

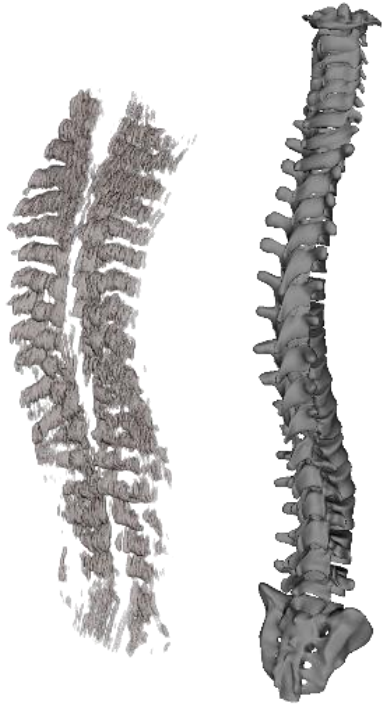

**Fig. 1.** Sample 3D reconstruction of a patient's spine and generic spine model.

## 4    Results and Discussion

### 4.1    Registration on ModelNet40

**Unseen Objects**. Table 1 presents registration performance on unseen objects. Figure 2 shows FPT's registration performance with rigid and non-rigid input transformations. FPT outperformed all other evaluated methods with respect to rotational error. FPT was comparable or superior to other methods with respect to translational error when presented with rigid and non-rigid input transformations. FPT was comparable to other learning-based methods with respect to inference time, and was over 50 times faster than CPD - an established and widely used iterative non-rigid registration method. Additionally, FPT performed consistently in both rigid and non-rigid registration, demonstrating its ability to effectively perform both types of registration.

**Table 1.** Registration performance of FPT and other methods on complete point sets of unseen ModelNet40 objects.

| Method | Transformation | Time (s) | RMSE (R) | RMSE (t) |
| --- | --- | --- | --- | --- |
| ICP (10 iter.) [24] | Rigid | 0.05 | 28.84 | 0.193 |
| PointNetLK [4] | Rigid | 0.14 | 14.47 | 0.045 |
| CorsNet [3] | Rigid | 0.08 | 16.24 | 0.012 |
| CPD [10] | Rigid | 5.94 | 8.29 | 0.049 |
| CPD [10] | Non-Rigid | 6.01 | 8.39 | 0.051 |
| FPT | Rigid | 0.08 | 5.01 | 0.015 |
| FPT | Non-Rigid | 0.08 | 5.18 | 0.032 |

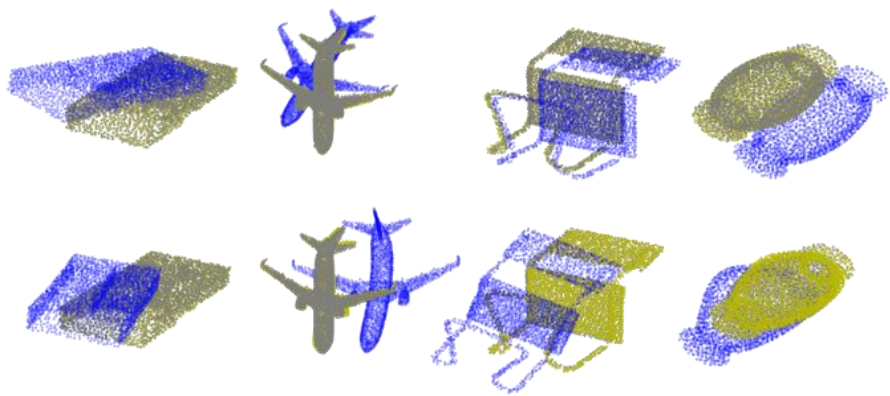

**Fig. 2.** Rigid (top) and non-rigid (bottom) registrations with FPT. Blue: source, yellow: target, gray: transformed source.

**Partial Registration with Unseen Objects**. Table 2 gives registration performance for 'partial-to-full' and 'partial-to-partial' registrations. Figure 3 shows FPT's 'partial-to-full' and 'partial-to-partial' non-rigid registration performance. Here, we see that FPT was comparable or superior to other learning-based methods. While PRNet maintains superior performance in both rotational and translation error, it is explicitly designed for partial registration and iteratively refines the predicted registration [4]. Furthermore, our use of a modified Chamfer Distance limits FPT's ability to perform partial registration as it relies on the existence of a one-to-one correspondence between point sets. When points are removed from the source or target point set, as in partial registration, a two-way Chamfer Distance will compute distances in a one-to-many manner, as some points will be distant to the other point set. However, this limitation may be alleviated through the formulation of a one-way Chamfer Distance loss for partial registration applications. As such, FPT's performance in partial registration must be further validated with other losses and training protocols in future work.

**Table 2.** Registration performance of FPT and other methods on partially occluded point sets of unseen ModelNet40 objects.

| Method | Transformation | Tranformation Type | RMSE (R) | RMSE (t) |
| --- | --- | --- | --- | --- |
| ICP (10 iter.) [24] | Rigid | Partial-to-partial | 32.40 | 0.279 |
| PointNetLK [1] | Rigid | Partial-to-partial | 16.58 | 0.048 |
| PRNet [4] | Rigid | Partial-to-partial | 3.20 | 0.016 |
| FPT | Rigid | Partial-to-partial | 6.97 | 0.063 |
| FPT | Non-Rigid | Partial-to-partial | 7.99 | 0.082 |
| FPT | Rigid | Partial-to-full | 5.79 | 0.053 |
| FPT | Non-Rigid | Partial-to-full | 6.34 | 0.068 |

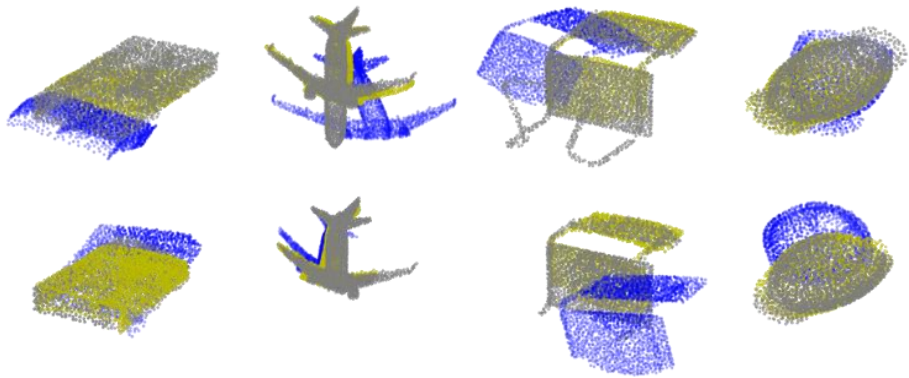

**Fig. 3.** Non-rigid 'partial-to-full' registrations at 25% occlusion of the source point set (top) and 'partial-to-partial' registrations at 25% occlusion of the source and target point sets (bottom) with FPT. Blue: source, yellow: target, gray: transformed source.

### 4.2 3D Spine Ultrasound Reconstruction Registration

TxAs were measured in the X-ray and deformed spine models to permit comparison between the clinical standard for scoliosis measurement and FPT's registration. The 13 spinal curvatures acquired from X-ray in our dataset measured between 6.4º and 11.5º. The maximum difference between TxA measurements from X-ray and deformed model was 2.3º. The average difference between TxA measurements from X-ray and deformed model was 1.3º. Figure 4 graphically demonstrates these results. Figure 5 illustrates an anterior-posterior and lateral visualization of a representative case from our dataset.

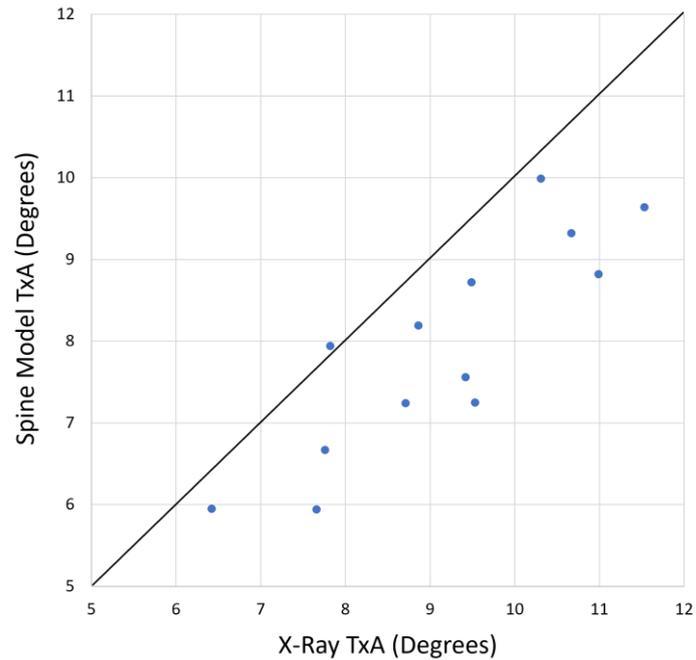

**Fig. 4.** Scatter plot of per-patient transverse process angles (TxA) measured from deformed spine models vs. X-ray based measurements.

In practice, curvatures larger than 5º are considered clinically significant. Additionally, curvatures from X-ray measurements may vary by up to 5º due to inter-observer variability and the time of day at which the images are acquired [26]. As such, for monitoring and measuring of scoliosis, an accuracy within 5–10º is considered acceptable for determining the next steps and course of care for a patient. While this proof-of-concept experiment is limited by the patient sample size, and the scale of the curvatures, given the availability of paired US imaging and corresponding X-rays, it is clear that the results we have presented are clinically acceptable. Without any fine-tuning, and having provided FPT with geometries which are external to the training set, FPT was able to register the spine models with an average error of less than 1.5º. Importantly, all measured differences fell within the 5º clinical error range, permitting a promising future use for the creation of accurate 3D visualizations that may be used for monitoring and measuring scoliosis without the need for error-prone manual registration processes, or the use of X-ray imaging and its associated ionizing radiation.

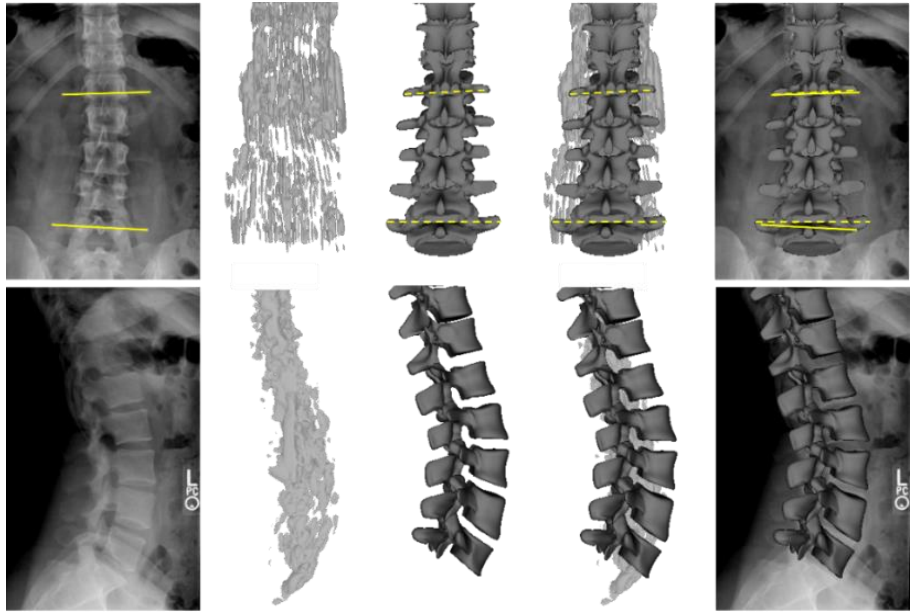

**Fig. 5.** Sample anterior-posterior (top) and lateral (bottom) visualization from patient data. Left to right: X-ray image, 3D reconstruction, deformed spine model, deformed spine model overlaid on 3D reconstruction, deformed spine model overlaid on X-ray image. Solid lines show measurements in X-ray images, dashed lines show measurements on deformed spine models.

## 5  Conclusion

Through evaluation with synthetic non-medical data and with US-based spine reconstructions, demonstrating a real-world clinical application, we demonstrated the effectiveness of the FPT architecture for point-set registration using deep neural networks. FPT is robust to deformation, and through our evaluation of atlas-based registration to US-based spine reconstructions, shown to be generalizable to geometries external to its training data domain. In other medical imaging problems where training data may be limited, FPT's generalizability may be of interest, given its ability to rapidly register point sets extracted from imaging acquired at different times or from different modalities. This further demonstrates FPT's utility as a generally-applicable method for learning-based non-rigid registration, representing significant progress for non-iterative, non-rigid point set registration without need for point correspondence.

**Acknowledgments.** This work is supported by the Wellcome/EPSRC Centre for Interventional and Surgical Sciences (203145Z/16/Z). Z.M.C. Baum is supported by the Natural Sciences and Engineering Research Council of Canada Postgraduate Scholarships-Doctoral Program, and the University College London Overseas and Graduate Research Scholarships.